\documentclass{article}
\usepackage[preprint,nonatbib]{neurips_2018}

\usepackage{longtable}

\usepackage{booktabs}

\usepackage{mathrsfs}
\usepackage{algorithm}
\usepackage{algorithmic}
\usepackage{wrapfig}
\usepackage{xcolor}

\usepackage{multirow}
\usepackage{lscape}
\usepackage{tabulary}
\usepackage{bbm}

\usepackage[utf8]{inputenc} 
\usepackage[T1]{fontenc}    
\usepackage{hyperref}       
\usepackage{url}            
\usepackage{booktabs}       
\usepackage{amsfonts}       
\usepackage{nicefrac}       
\usepackage{microtype}      

\usepackage{amsmath}
\usepackage{algorithm, algorithmic}
\usepackage{amsfonts, amssymb}
\usepackage{multirow}
\usepackage{mathtools}
\usepackage{arydshln}
\usepackage{refcount}
\usepackage{listings}
\usepackage{hyperref}

\newtheorem{theorem}{Theorem}[section]

\DeclareMathOperator*{\argmin}{arg\,min}

\title{Greedy UnMixing for Q-Learning in Multi-Agent Reinforcement Learning}

\author{%
  Chapman Siu \\
  Faculty of Engineering and Information Technology \\
  University of Technology Sydney, Australia \\
  \texttt{chpmn.siu@gmail.com} \\
  \And
   Jason Traish \\
  Faculty of Engineering and Information Technology \\
  University of Technology Sydney, Australia \\
  \texttt{Jason.Traish@uts.edu.au} \\
  \AND
  Richard Yi Da Xu \\
  Faculty of Engineering and Information Technology \\
  University of Technology Sydney, Australia \\
  \texttt{YiDa.Xu@uts.edu.au} \\
}

\begin{document}

\maketitle

\begin{abstract}
This paper introduces Greedy UnMix (GUM) for cooperative multi-agent reinforcement learning (MARL). Greedy UnMix aims to avoid scenarios where MARL methods fail due to overestimation of values as part of the large joint state-action space. It aims to address this through a conservative Q-learning approach through restricting the state-marginal in the dataset to avoid unobserved joint state action spaces, whilst concurrently attempting to unmix or simplify the problem space under the centralized training with decentralized execution paradigm. We demonstrate the adherence to Q-function lower bounds in the Q-learning for MARL scenarios, and demonstrate superior performance to existing Q-learning MARL approaches as well as more general MARL algorithms over a set of benchmark MARL tasks, despite its relative simplicity compared with state-of-the-art approaches. 
\end{abstract}

\section{Introduction}


In multi-agent reinforcement learning, the nature of the joint state action space leads to complexity in estimation, as the size of the joint action spaces grow exponentially with the number of agents. In order to constrain and learn policies effectively, we leverage approaches within offline reinforcement learning, which is used to avoid overestimation of values induced by distribution shift between the learned policy and experiences of the multiple agents. In order to scale out reinforcement learning, especially in adherence to constraints on agent observability and communication in multi-agent environments, {\em decentralized} policies are used so that agents may only act on their local observations. Previous work, and this work, leverage {\em centralized training with decentralized execution} (CTDE) \cite{qmix,lica,qtran,vdn,maddpg}, where independent agents can access additional state information, (including other agents observation and actions) that is unavailable during policy inference. 

However the key question remains - is there a way to determine the usefulness of this additional state information? Is there a way to effectively ``unmix'' this additional state information, removing the need to learn these interactions in the first place? We propose a method for concurrently simplifying the problem space, and learning a conservative estimate for the joint value function, which allows for practical simplification of the joint state action space in MARL settings. We demonstrate this idea in a novel Q-learning algorithm called {\em Greedy UnMix} (GUM). First, GUM leverages conservative estimation attempts to avoid poor performance due to issues with bootstrapping out of distribution actions \cite{bear,cql} and overfitting, which arises from erroneous optimistic value function estimates. In order to achieve this, conservative estimates of the value function could be used to address the overestimation problem. In our experiments, we demonstrate empirically the importance of providing a conservative estimate on the {\em decentralized} policies and providing a conservative estimate on the {\em mixer} network is required in order to avoid over optimistic estimations. Furthermore, adaptive control of the regularization effect in the conservative estimate between the mixing and independent policies is necessary in order to effectively unmix and ensemble the Q-learning estimates to stabilize learning. We evaluate GUM over several benchmark multi-agent environments, and find that it achieves superior results to other Q-learning MARL approaches, and is competitive with state of the art MARL approaches, whilst maintaining the simplicity of Q-learning MARL approaches.

\section{Preliminaries}

\subsubsection{Dec-POMDP:}

We assume a Dec-POMDP for $n$ agents $\langle S, U, P, r, Z, O, n, \gamma \rangle$, where $S$ is the state space of the environment \cite{decpomdp}. At each timestep, $t=1, 2, 3, \dots$, every agent $i \in \mathcal{A} \equiv \{1, \dots, n \}$, which chooses an action $u^i \in U$ which forms the joint action $\mathbf{u} \in \mathbf{U} \equiv U^n$. At each time step $t$, the next state $s \in S$ is drawn from state transition function $P(. | s, \mathbf{u})$. A transition yields a collaborative reward $r(s, \mathbf{u})$, and $\gamma \in [0, 1)$ denotes the discount factor. We consider {\it partially observable} settings, where each agent does not have full state, but samples observations $z \in Z$ according to observation function $O(s, i): S \times \mathcal{A} \rightarrow Z$. We denote the action observation history for an agent $i$ to be $\tau^i \in T \equiv (Z \times U)^*$, on which it conditions its stochastic policy $\pi^i(u^i | \tau^i): T \times U \rightarrow [0, 1]$.  The joint policy $\pi$ is based on the {\it action-value function}: $Q^\pi(s_t, \mathbf{u}_t) = \mathbb{E}_{s_{t+1:\infty}, \mathbf{u}_{t+1:\infty}}[R_t|s_t, \mathbf{u}_t]$, where $R_t = \sum_{i=0}^\infty \gamma^i r_{t+i}$ which is the {\it discounted reward}. The goal of the problem is to find the optimal action value function $Q^*$. Q-learning methods train the Q-function by iteratively applying the Bellman optimality operator $\mathcal{B}^*Q = r(s, \mathbf{u}) + \gamma \mathbb{E}_{s^\prime \sim P(s^\prime \vert s, \mathbf{u})}[\max_{a^\prime} Q(s^\prime, \mathbf{u}^\prime)]$. Actor-critic methods alternate between computing $Q^\pi$ through iterating the Bellman operator, $\mathcal{B}^\pi Q = r + \gamma P^\pi Q$, where $P^\pi$ is the transition matrix coupled with the policy, and improving the policy $\pi(u \vert s)$ by updating it towards actions that maximize the expected Q-value.


\subsubsection{Centralized Training Decentralized Execution (CTDE):}

In this setting we further consider the scenario where training is centralized, and execution is decentralized; that is, during training, the learning algorithm has access to the action-observation histories of all agents and the full state, however each agent can only condition on its own local action-observation history $\tau^i$ during decentralized execution.




\subsubsection{Overcoming Bootstrap Accumulation Error in Reinforcement Learning:} Overcoming bootstrap accumulation error is a key focus in offline learning (or batch reinforcement learning), whereby the temporal difference leverages a bootstrapped target network, which leads to overly optimistic Q function estimations when attempting to extrapolate unseen or out of distribution actions. These approaches leverage large datasets of experience in order to train policies without active interaction with the environment.  This differs from imitation learning, which assumes optimal data, as the dataset of experiences may be sub-optimal or even random! Approaches to resolve this avoid actions which lie outside the observed data distribution \cite{bear,bcq}, or modeling the distribution of actions which can reduce overestimation through providing a lower-bound estimate \cite{cql,fujimoto2019benchmarking,qrdqn}. Ensemble approaches have also been explored for stabilising Q-functions \cite{bcq,rem}. 


\subsubsection{Multi-Agent Reinforcement Learning:} Within the MARL approaches which focus on the CTDE paradigm, one approach is independent Q-learning \cite{iql} whereby each agent consists of its own Q-learning policy. Approaches which leverage joint state space are mostly composed through the construction of the Independent Global Max problem \cite{qtran}, whereby a Q-learner with parameter sharing is learned for each agent, and then composed with a mixing function. This alters the Q-learning problem with an objective to leverage a mixer function. In value decomposition approach \cite{vdn}, each individual value function is combined in a linear fashion; QMIX \cite{qmix} combines the value functions through a neural network with monotonic constraints, and QTRAN demonstrates how arbitrary value factorization can occur, leading to computationally intractable constraints in learning the joint state action spaces in the MARL setting. 

Another approach to MARL problems is through a policy gradient framework or actor critic framework. These typically operate through each agent having its own policy, but the critic may be composed in different ways. One approach is for each agent to have its own policy and critic network, in which experience is shared across multiple agents \cite{seac}; in the deterministic policy gradient setting, the critic is composed through constructing a joint state-action space \cite{maddpg}. The use of a mixer network similar to QMIX has been explored in which a mixer network is used as a critic with adaptive entropy regularization \cite{lica}.

\section{Methodology}

In this section we develop Greedy UnMixing (GUM) algorithm, such that the expected value of the policy under the learned Q-function lower-bounds its true value. Naively we could use independent Q-learning \cite{iql} approach and apply it to conservative Q-learning approach. We further expand this through applying lower-bound to the joint Q-learning mixer function in the multi-agent setting and provide an approach to unmix the joint Q-learning and independent Q-learning procedure.


\subsection{Independent Conservative Q-Learning}

\label{ciql}

Conservative Q-learning can be used in off-policy scenario, and is formulated through the addition of a conservative policy regularizer which uses quantile regression Q-learning as the base algorithm so that a distribution of values can be learned to calculate the action distribution \cite{qrdqn}. The CQL framework can be applied in MARL framework in the same way as how DQN is applied in IQL algorithm - that is, each agent learns its own independent Conservative Q-learning policy conditioned on its own partial observation \cite{iql}. It operates through the addition of a penalty which minimizes the expected Q-value under a particular distribution of state-action pairs $\mu$ (can be constructed through the restricting $\mu$ to match the state-marginal in the dataset), which can be used to match the state-marginal in the dataset, denoted by $\mathbb{E}_{\tau \sim \mathcal{D}, u \sim \mu} [Q(\tau, u)]$. We define the dataset by $\mathcal{D}$, so that on all states let $\hat{\beta}(u \vert \tau) := \frac{\sum_{\tau, u \in \mathcal{D}}\mathbbm{1}[\tau=\tau, u = u]}{\sum_{\tau \in \mathcal{D}}\mathbbm{1}[u=u]}$ denote the empirical behavior policy at the partially observed state. This is further improved through the addition of Q-value {\em maximization} term $\mathbb{E}_{\tau \sim \mathcal{D}, u \sim \mu} [Q(\tau, u)] - \mathbb{E}_{\tau \sim \mathcal{D},u \sim \hat{\beta}} [Q(\tau, u)]$, with a trade-off factor for the penalty $\alpha$. This iterative update {\em per each individual agent} is shown below based on the CQL algorithm \cite{cql}: 



\begin{align}
\hat{Q}^{\pi}_{\text{CQL}} \leftarrow \underset{Q}{\argmin} \text{ } \underset{\text{off-policy CQL regularizer}}{\underbrace{ \alpha \cdot \left( \mathbb{E}_{\tau \sim \mathcal{D}, u \sim \mu} [Q(\tau, u)] - \mathbb{E}_{\tau \sim \mathcal{D},u \sim \hat{\beta}} [Q(\tau, u)] \right) }} \nonumber \\
+ \frac{1}{2}\underset{\text{standard TD error}}{\underbrace{\mathbb{E}_{\tau, u, \tau^\prime \sim \mathcal{D}}\left[(r(\tau, u) + \gamma \mathbb{E}_\pi [\bar{Q}(\tau^\prime, u^\prime)] - Q(\tau, u))^2 \right]}} \label{eq:icql}
\end{align}

As each agent leverages CQL {\em independently}, then each agent conservatively estimates Q-values independently from each other, which allows for theoretical guarantees to hold. This leads to the natural question; under what conditions the appropriate value function factorization would still retain conservative Q-value guarantees which will provide the lower-bounds of the policy?

\subsection{Conservative Monotonic Value Function Factorization}

\label{cqmix}

QMIX leverages monotonic value function factorization, in which the mixer hypernetwork $Q_\text{mixer}$ consists of monotonic transformations of each agent Q function, $Q_1, Q_2, \dots, Q_n$. The naive approach is to regularize with respect to the whole hypernetwork, in a similar fashion as the Conservative IQL approach in the previous Section \ref{ciql}, the $Q_{\text{mixer}}$ iterative update, where the full joint state is $s$, and the joint action is $\mathbf{u}$:


\begin{align}
\hat{Q}^{\pi}_{\text{QMIX}} \leftarrow \underset{Q}{\argmin} \text{ } \underset{\text{off-policy CQL regularizer}}{\underbrace{\alpha \left( \mathbb{E}_{s \sim \mathcal{D}, \mathbf{u} \sim \{\mathbf{\mu_i}\}_{i=1}^n} [Q_\text{mixer}(s, \mathbf{u})] - \mathbb{E}_{s \sim \mathcal{D},\mathbf{u} \sim \{\hat{\beta_i}\}_{i=1}^n} [Q_\text{mixer}(s, \mathbf{u})] \right) }}  \nonumber \\
+ \frac{1}{2}\underset{\text{standard TD error}}{\underbrace{\mathbb{E}_{s, u, s^\prime \sim \mathcal{D}}\left[(r(s, \mathbf{u}) + \gamma \mathbb{E}_\pi [\bar{Q}_\text{mixer}(s^\prime, \mathbf{u}^\prime)] - Q_\text{mixer}(s, \mathbf{u}))^2 \right]}} 
\label{eq:qmix}
\end{align}

This approach would regularize the weights in both the mixer hypernetwork and the individual agent policy networks. 

However, this is not entirely necessary, as regularizing only the individual agent policies would also provide a lower bound on the Q function estimation; the effect of this will be regularising the policy weights but not the weights in the monotonic hypernetwork. As the mixer network is not used directly in the execution of the policy, it only affects the weights updating in the training step. We will explore the effect of regularizing the mixer network, or the policy network, or both concurrently in the ablation studies. 

In consideration of the independent global max formulation, to guarantee that the condition holds, we similarly apply the CQL regularizer to the $Q_\text{mixer}$ function. Replacing $Q$ with the mixer function would adhere to the theoretical considerations due to the monotonic nature of QMIX. 



\subsection{Greedy UnMixing}


\label{gum}

The construction of Greedy UnMixing (GUM) takes into account the manner in which conservative QMIX is constructed, with both regularizer on the mixer network and individual network. In determining the updates of the mixer network or the individual policy networks, we consider the nature in which they are updated. Can we bias the weights towards considering individual information only and {\em purposely} neglect the coordination aspect of the mixer network, thereby ``unmixing'' the network? 

\begin{figure}[h]
  \begin{center}
    \includegraphics[width=0.22\textwidth]{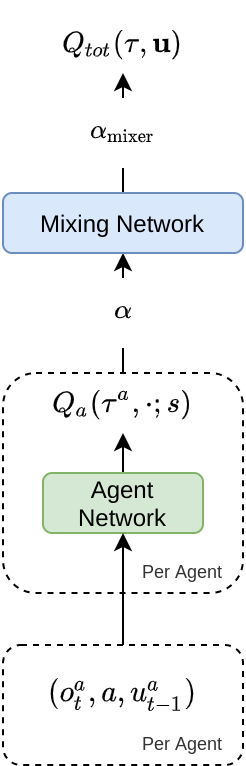}
  \end{center}
\caption{Architecture for the GUM algorithm. The model is implicitly unmixed between the independent policies and cooperative training used in the mixer network, through weighting the regularizer strength through a learned $\alpha$ and $\alpha_\text{mixer}$. This allows for learning the assigning learning weights to the independent policy and mixer network in an appropriate manner to avoid overestimation.}
\label{gum-orig}
\end{figure}

In order to unmix, we use a learnable weight to control the regularization strength for the independent and mixer networks. Intuitively, the ``better'' the estimate the stronger the strength of the regularizer. In order to construct this, we apply a soft constraint on the regularizer trade-off factor, such that the larger the relative regularizer strength, the smaller its corresponding component has on the neural network weights update. To simplify notation, we represent the regularizer term from Equation \ref{eq:icql} to be $\mathcal{R}$ and regularizer term from Equation \ref{eq:qmix} to be $\mathcal{R}_{\text{mixer}}$, with the trade-off factors $\alpha$ and $\alpha_{\text{mixer}}$ respectively and is diagrammatically shown in Figure \ref{gum-orig}. Then the iterative update is provided as

\begin{align}
\hat{Q}^{\pi}_{\text{GUM}} \leftarrow \alpha\mathcal{R} + \alpha_{\text{mixer}}\mathcal{R}_{\text{mixer}} \nonumber \\
+ \frac{1}{2}{\mathbb{E}_{s, \mathbf{u}, s^\prime \sim \mathcal{D}}\left[(r(s, \mathbf{u}) + \gamma \mathbb{E}_\pi [\bar{Q}_\text{mixer}(s^\prime, \mathbf{u}^\prime)] - Q_\text{mixer}(s, \mathbf{u}))^2 \right]}
\label{eqn:objective}
\end{align}


%



We explored GUM, in which regularizer trade-off weights were used to implicitly to unmix the cooperative and independent Q functions. Next we explore the effects for explicitly unmixing the two components through an ensemble interpretation of GUM as a neural network architecture selection problem.

\subsection{Ensemble Learning Interpretation of Greedy UnMix}

\label{ensemblegum}


The construction and choice of neural network architectures have been explored as a boosting problem with learning guarantees. AdaNet framework explores neural architecture selection through iteratively proposing candidate networks with hard selection proposals \cite{adanet}, or soft selection proposals through application of a learnable trade-off factor \cite{siu2019residual}. Through the learnable trade-off $\alpha$, the GUM algorithm can be interpreted as a neural network architecture selection problem at training time between the mixer network and individual agent networks. This ensemble can be applied implicitly, as suggested in Section \ref{gum} or explicitly as a stacking ensemble \cite{Wolpert1992} of the individual agent policies and the mixer function. Rather than leveraging a learned parameter for neural network selection problem, an alternative approach is to simply take the minimum ensemble as the solution, which leads to a hard assignment in the choice of the Q function ensemble rather than a soft assignment. This approach is a variety of ensemble Q-learning approaches \cite{ddpg,bcq,bear}, which is used to stabilize training in the context of homogeneous target networks, as opposed to this scenario where it is between the individual agent policies and the mixer network. Then the Adanet mixer is a linear combination of the individual agent Q function, and the monotonic function factorization, $Q_{\text{adanet}} = \alpha [Q_1, \dots, Q_n] + \alpha_{\text{mixer}} Q_{\text{mixer}}$, where $\alpha$ and $\alpha_{\text{mixer}}$ are learnable parameters by gradient ascent. The iterative update is given by

\begin{align}
\label{eqn:adanet}
\hat{Q}^{\pi}_{\text{GUM-adanet}} \leftarrow {\mathbb{E}_{s, \mathbf{u}, s^\prime \sim \mathcal{D}}\left[(r(s, \mathbf{u}) + \gamma \mathbb{E}_\pi [\bar{Q}_\text{adanet}(s^\prime, \mathbf{u}^\prime)] - Q_\text{adanet}(s, \mathbf{u}))^2 \right]}
\end{align}

\begin{algorithm}[H]
\caption{Greedy UnMixing Algorithm}
\label{alg:gum}
\begin{algorithmic}[1]

\STATE Initialize Q-function for each agent ${Q_i}_{i=1}^n$, and the mixer hypernetwork $Q_\text{mixer}$, Lagrange multiplier $\alpha, \alpha_{\text{mixer}}$.
\FOR{step $t$ in $\{1, \dots, N\}$}
    \STATE Train ${Q_i}_{i=1}^n$ and the mixing network, $Q_\text{mixer}$ functions using gradient step using objective from Equation \ref{eqn:objective} for implicit approach or Equation \ref{eqn:adanet} for explicit approach.
    \STATE Update Lagrange multiplier $\alpha, \alpha_\text{mixer}$, with soft constraints inversely proportional to regularizer strength. 
\ENDFOR
\STATE Evaluate policies using ${Q_i}_{i=1}^n$
\end{algorithmic}
\end{algorithm}

\begin{figure}[h]
  \begin{center}
    \includegraphics[width=0.55\textwidth]{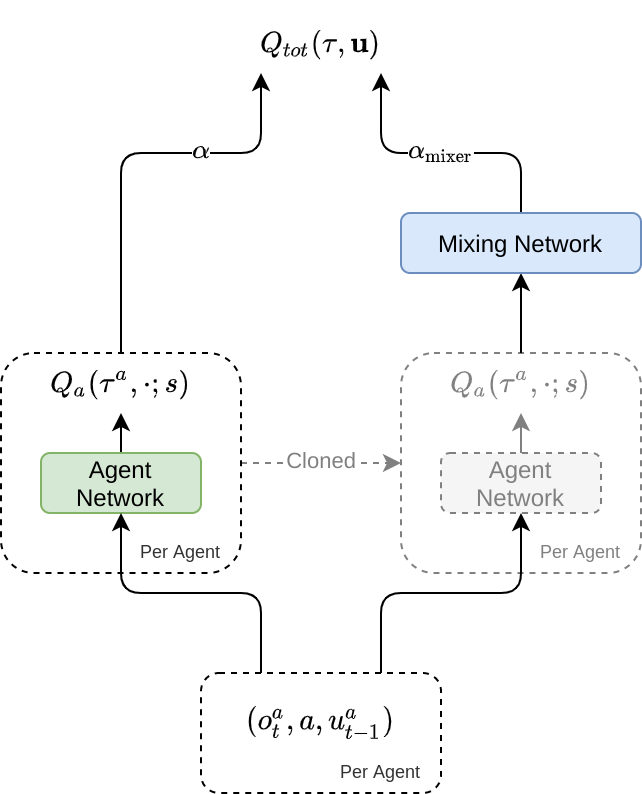}
  \end{center}
\caption{Architecture for the GUM algorithm can be interpreted as an ensemble of neural network modules in an AdaBoost setting with the candidate network being included is the mixer network. This interpretation allows for explicit unmixing between the independent policies and the cooperative training used in the mixer network.}
\label{gum-adanet}
\end{figure}

Another interpretation of the AdaNet neural network selection setting, is that the base learner is the independent Q-learning policies, which are then boosted (ensemble) from the subsequent mixer network. This interpretation is shown in Figure \ref{gum-adanet} and also addresses the overestimation concerns in reinforcement learning, through penalizing the additional weight inversely relative to the regularizer strength. There have been several approaches leveraging ensemble of Q functions which assume independence and their analysis results would not apply here either \cite{bcq,bear}. In the boosting setting, our approach can be interpreted as a variation of AdaNet \cite{adanet}, in the bushy construction, whereby the independent Q-learning model is the base policy, and the mixer is the boosted network module. Rather than having hard update or stopping rule, we leverage a soft choice as a shrinkage parameter \cite{siu2019residual}. In this setting, the networks are explicitly combined rather than implicitly through the regularizer.

We present the two variations of GUM, one which allows for implicit unmixing through learning the trade-off factors through an inverse weighting scheme, and the other approach in which the combination of the mixer and independent policies are explicitly learned, summarized in algorithm \ref{alg:gum}.

\subsubsection{Theoretical Considerations of GUM.}


GUM is an extension of conservative Q-learning, and in the implicit regularizer formulation, extends the theoretical results from \cite{cql1,cql2,cql} to multi-agent setting which are related to prior work on safe policy improvement \cite{laroche19,laroche16}. 

\begin{theorem}
{\normalfont (Equation \ref{eqn:objective} results in a lower bound estimate of the true policy)} The value of the an independent agent's policy under the $Q_{\text{GUM}}$ function from Equation \ref{eqn:objective},  $\hat{V}^\pi(s) = \mathbb{E}_\pi [\hat{Q}_{\text{GUM}}(s, u)]$, lower-bounds the true value of the policy $V^\pi(s) = \mathbb{E}_\pi [{Q}_{\text{GUM}}(s, u)]$, when $\mu = \pi$ for any $\alpha > 0$ in the tabular setting.
\label{theorem:lb}
\end{theorem}

{\em Proof:} We can set the modified objective in Equation \ref{eqn:objective} and compute the Q-function update induced in the exact tabular setting. Let $k \in \mathbb{N}$ denote an iteration of policy evaluation, so that at iteration $k$ the objective - Equation \ref{eqn:objective} - is optimized. To demonstrate that Equation \ref{eqn:objective} results in a lower bound estimate of the true policy, we consider the value function update:

\begin{align}
    \hat{V}^{k+1}(\tau) &:= \mathbb{E}_{\mathbf{u} \sim \{\pi_i(u \vert \tau)\}_{i=1}^n}\left[\hat{Q}^{k+1}(s, \mathbf{u} )\right] \nonumber \\
    &= \mathcal{B}^* \hat{V}^k(\tau) - \alpha \left(\sum_{i=1}^n \mathbb{E}_{u \sim \pi_i(u \vert \tau_i)}\left[\frac{\mu_i(u \vert \tau_i)}{\beta(u \vert \tau_i)} -1\right] \right) \nonumber \\
    &\qquad - \alpha_{\text{mixer}} \mathbb{E}_{\mathbf{u} \sim \pi_{\text{mixer}}(\mathbf{u} \vert s)}\left[\frac{\mu_{\text{mixer}}(\mathbf{u} \vert s)}{\beta(\mathbf{u} \vert s)} -1\right]  \nonumber
\end{align}

For any particular agent, if $\mu_i = \pi_i$, we first consider considering (dropping the $i$ for brevity) $\mathbb{E}_{u \sim \pi(u \vert \tau)}\left[\frac{\mu(u \vert \tau)}{\beta(u \vert \tau)} -1\right]$, we have the following derivation which follows the result from CQL Theorem 3.2 \cite{cql}:

\begin{align*}
    \mathbb{E}_{u \sim \pi(u \vert \tau)}\left[\hat{Q}^{k+1}(\tau, u)\right] & = \sum_u \pi(u \vert \tau) \left[\frac{\mu(u \vert \tau)}{\beta(u \vert \tau)} -1\right] \\
    & = \sum_u (\pi(u \vert \tau) - \beta(u \vert \tau) + \beta(u \vert \tau)) \left[\frac{\mu(u \vert \tau)}{\beta(u \vert \tau)} -1\right] \\
    & = \sum_u (\pi(u \vert \tau) - \beta(u \vert \tau))\left[ \frac{\pi(u \vert \tau) - \beta(u \vert \tau)}{\beta(u \vert \tau)}\right] \\
    &\qquad + \sum_u \beta(u \vert \tau) \left[\frac{\mu(u \vert \tau)}{\beta(u \vert \tau)} -1\right] \\
    & = \sum_u \left[ \frac{(\pi(u \vert \tau) - \beta(u \vert \tau))^2}{\beta(u \vert \tau)}\right] + \sum_u \mu(u \vert \tau) - \sum_u \beta(u \vert \tau) \\
    & = \sum_u \left[ \frac{(\pi(u \vert \tau) - \beta(u \vert \tau))^2}{\beta(u \vert \tau)}\right] + 0\\ 
    & \geq 0
\end{align*}

Then $\sum_{i=1}^n \mathbb{E}_{u \sim \pi_i(u \vert \tau_i)}\left[\frac{\mu_i(u \vert \tau_i)}{\beta(u \vert \tau_i)} -1\right]$ is non-negative, as each independent agent regularizer is additive. A similar argument is applied for the joint state action mixer component $\mathbb{E}_{\mathbf{u} \sim \pi_{\text{mixer}}(\mathbf{u} \vert s)}\left[\frac{\mu_{\text{mixer}}(\mathbf{u} \vert s)}{\beta(\mathbf{u} \vert s)} -1\right]$.

\begin{align*}
    \mathbb{E}_{\mathbf{u} \sim \pi_{\text{mixer}}(\mathbf{u} \vert s)}\left[\frac{\mu_{\text{mixer}}(\mathbf{u} \vert s)}{\beta(\mathbf{u} \vert s)} -1\right] & = \sum_{\mathbf{u}  \in \mathbf{U}} \pi_{\text{mixer}}(\mathbf{u} \vert s) \left[\frac{\mu_{\text{mixer}}(\mathbf{u} \vert s)}{\beta(\mathbf{u} \vert s)} -1\right]\\
    &= \sum_{\mathbf{u}  \in \mathbf{U}} \Big(\left(\pi_{\text{mixer}}(\mathbf{u} \vert s) - \beta(\mathbf{u} \vert s) + \beta(\mathbf{u} \vert s) \right) \\
    &\qquad \times \left[\frac{\mu_{\text{mixer}}(\mathbf{u} \vert s)}{\beta(\mathbf{u} \vert s)} -1\right]\Big) \\
    &= \sum_{\mathbf{u}  \in \mathbf{U}} \left[ \frac{(\pi_{\text{mixer}}(\mathbf{u} \vert s)- \beta(\mathbf{u} \vert s))^2}{\beta(\mathbf{u} \vert s)} \right] \\
    &\qquad + \sum_{\mathbf{u}  \in \mathbf{U}}\mu_{\text{mixer}}(\mathbf{u} \vert s)  +  \sum_{\mathbf{u}  \in \mathbf{U}} \beta(\mathbf{u} \vert s) \\
    &\geq 0
\end{align*}

As $\mathbb{E}_{u \sim \pi_i(u \vert \tau_i)}\left[\frac{\mu_i(u \vert \tau_i)}{\beta(u \vert \tau_i)} -1\right], \forall i$ and $\mathbb{E}_{\mathbf{u} \sim \pi_{\text{mixer}}(\mathbf{u} \vert s)}\left[\frac{\mu_{\text{mixer}}(\mathbf{u} \vert s)}{\beta(\mathbf{u} \vert s)} -1\right]$ is non-negative, then each value iterate of GUM algorithm incurs some underestimation $\hat{V}^{k+1}(s) \leq \mathcal{B}^* \hat{V}^k(s)$. This result can be generalized to incorporate sampling error in the non-tabular setting in the actor-critic setting based on prior work \cite{cql}.

\section{Experiments}

\begin{figure}[ht]
\centering
\begin{tabular}{cccc}
\includegraphics[width=0.2\textwidth,height=0.2\textwidth]{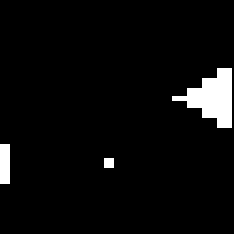} & \includegraphics[width=0.2\textwidth,height=0.2\textwidth]{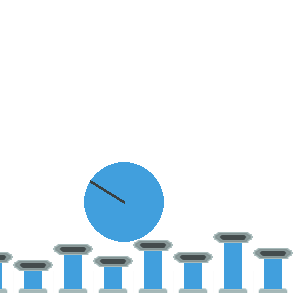} &
\includegraphics[width=0.2\textwidth,height=0.2\textwidth]{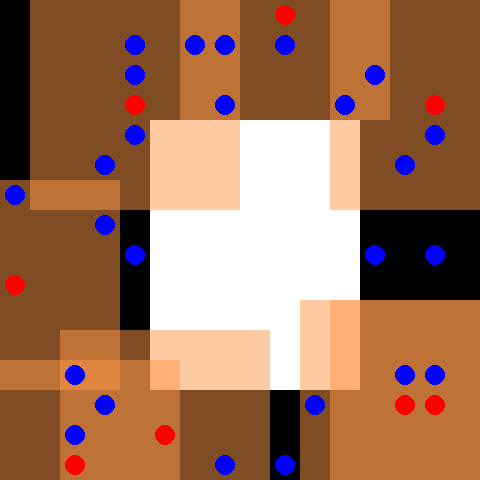} &
\includegraphics[width=0.2\textwidth,height=0.2\textwidth]{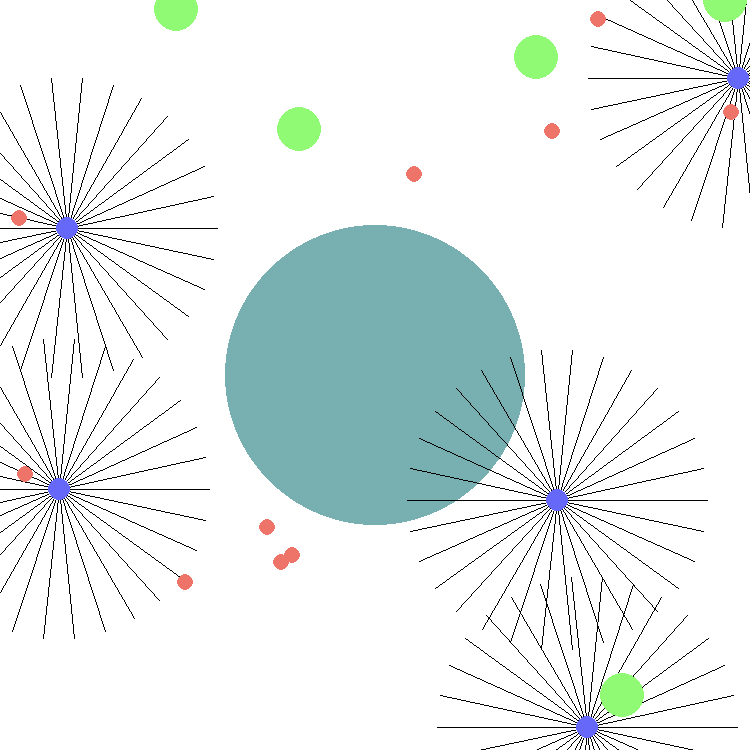}
\end{tabular}
\caption{Left to right: Pong, Pistonball, Pursuit, Waterworld environments.}
\label{fig:pettingzoo}
\end{figure}

In this section we describe experiments we use to empirically justify our algorithm and approach. We use several reinforcement learning benchmarks \cite{terry2020pettingzoo,maddpg,gupta2017cooperative}, shown in Figure \ref{fig:pettingzoo}.

{\em Cooperative Pong} is a multi-agent game of pong from ``Butterfly'' environments \cite{terry2020pettingzoo}, where the goal is for both agents (paddles) to keep the ball in play. The game is over if the ball goes out of bounds from either the left or right edge of the screen. In this setting the action space is discrete, and the paddles can either move up or down. To make learning more challenging the right paddle is tiered cake-shaped by default, and the observations of each agent are its own half of the screen. The agents receive a positive, fixed, combined reward based on successful completion of a game frame, otherwise a negative reward is given, and the game ends. 

{\em Pistonball} is a physics based cooperative game from ``Butterfly'' environments \cite{terry2020pettingzoo} where the goal is move the a ball to the left wall of the game border by activating vertically moving pistons. Each piston agent's observation is an RGB image of is local surrounding area. The agents have a discrete action space, and can choose to stay still, move down or move up. Pistons must learn highly coordinated emergent behavior to achieve an optimal policy for the environment. Each agent gets a reward based on the overall movement of the ball, and how much the ball moved in a left-wards direction if it was in the agent's local observation. 

{\em Pursuit} is an environment from the ``SISL'' set of environments \cite{gupta2017cooperative}, where our agents are control of pursuer agents, which aim to capture randomly controlled evader agents. Evader agents are only removed from the environment when pursuers fully surround an evader. Agents receive a small reward when they touch an evader, and a large reward when they successfully surround and remove an evader agent. In this scenario, each pursuer agent observes a local RGB grid, and operate in a discrete action space. 

{\em Waterworld} is a simulation of archea trying to navigate and surviving their environment from the ``SISL'' set of environments \cite{gupta2017cooperative}. The goal of the agent is to control the archea to simultaneously avoid poison and find food. The original environment consists of a continuous action space, indicating their horizontal and vertical thrust within the environment. We instead modify this to a discrete action space, in order to assess Q-learning approaches. Within this environment, the agent's observation is based on range limited sensors to detect neighboring entities, indicated by black lines in the right-most image in Figure \ref{fig:pettingzoo}. 

{\em Reference} and {\em Spread} are environments consisting of agents and landmarks from the ``MPE'' set of environments \cite{maddpg}. In {\em Reference}, each agent wants to get closer to their target landmark, which is known only by other agents. The agents are rewarded by the distance to their target landmark. In {\em Spread} Agents must learn to cover each landmark without colliding with other agents. The agents are rewarded based on how far the closest agent is to each landmark, and are penalized if they collide with other agents. For both environments, the agent's observation space consists of distance to every landmark, and the action space consists of the movement action. In addition the {\em Reference} environment has an additional actions to communicate with other agents.

\subsection{Architecture Overview and Implementation Details}

Across all environments, we consistently preprocessed the inputs as suggested in the PettingZoo benchmarks \cite{terry2020pettingzoo}. We evaluated on 1000 evaluation episode rollouts (separate from the train distributions) every training iteration and used the average score and variance for the plots and tables, shown in Figure \ref{resall}

\subsection{Hyper-parameters}
\begin{table*}[h]
\centering
\begin{tabular}{c|c}
\hline
\textbf{Hyperparameters}    & \textbf{Value} \\ \hline
Policy Hidden Sizes         & {[}64, 64, 64{]} \\
Mixer Hidden Sizes         & {[}32, 32, 32{]} \\
Policy Hidden Activation    & ReLU           \\
Target Network $\tau$          & 0.005          \\
Learning Rate               & 0.0003         \\
Batch Size                  & 256            \\
Replay Buffer Size          & 1000000        \\
Number of pretraining steps & 1000           \\
Steps per Iteration         & 1000           \\
Discount                    & 0.99           \\
Reward Scale                & 1              \\ \hline
\end{tabular}
\caption{Hyper-parameters used for MARL experiments}
\label{app:tab:hyper}
\end{table*}
\label{ref:hyper}

The policies leverage the default hyper-parameters based on the official implementation of LICA, QMIX, IQL and their accompanying code based on the pymarl library \cite{qmix,lica,qtran}. SEAC was re-implemented in rlkit framework with the same hyperparameters proposed in their paper, but using Soft Actor-Critic instead of A2C. Similarly, MADDPG extended the rlkit implementation of DDPG to support multi-agent setting. In order to facilitate discrete action spaces gumbel softmax \cite{gumbel_softmax1,gumbel_softmax2} is used as proposed by the approach taken by by SEAC \cite{seac}. Similar to the usage in the pymarl library, parameter sharing was used in all models in order to improve the general performance of the agents. 

Our particular implementation of GUM leverages the existing approach of Conservative Q-learning, which regularizes through an entropy interpretation of the off-policy CQL regularizer, which is framed as $$\underset{Q}{\min} \text{ } \mathbb{E}_{\tau \sim \mathcal{D}} \left[ \log \sum_{\mathbf{u}} \exp(Q(\tau, u)) - \mathbb{E}_{u \sim \mu}[Q(\tau, u)]\right]$$  where the distribution over data is modeled after the previous policy, as suggested and used in the official CQL implementation \cite{cql}.

\subsection{Ablations}

\begin{figure}

  \begin{center}
\begin{tabular}{cc}
\includegraphics[width=0.45\textwidth]{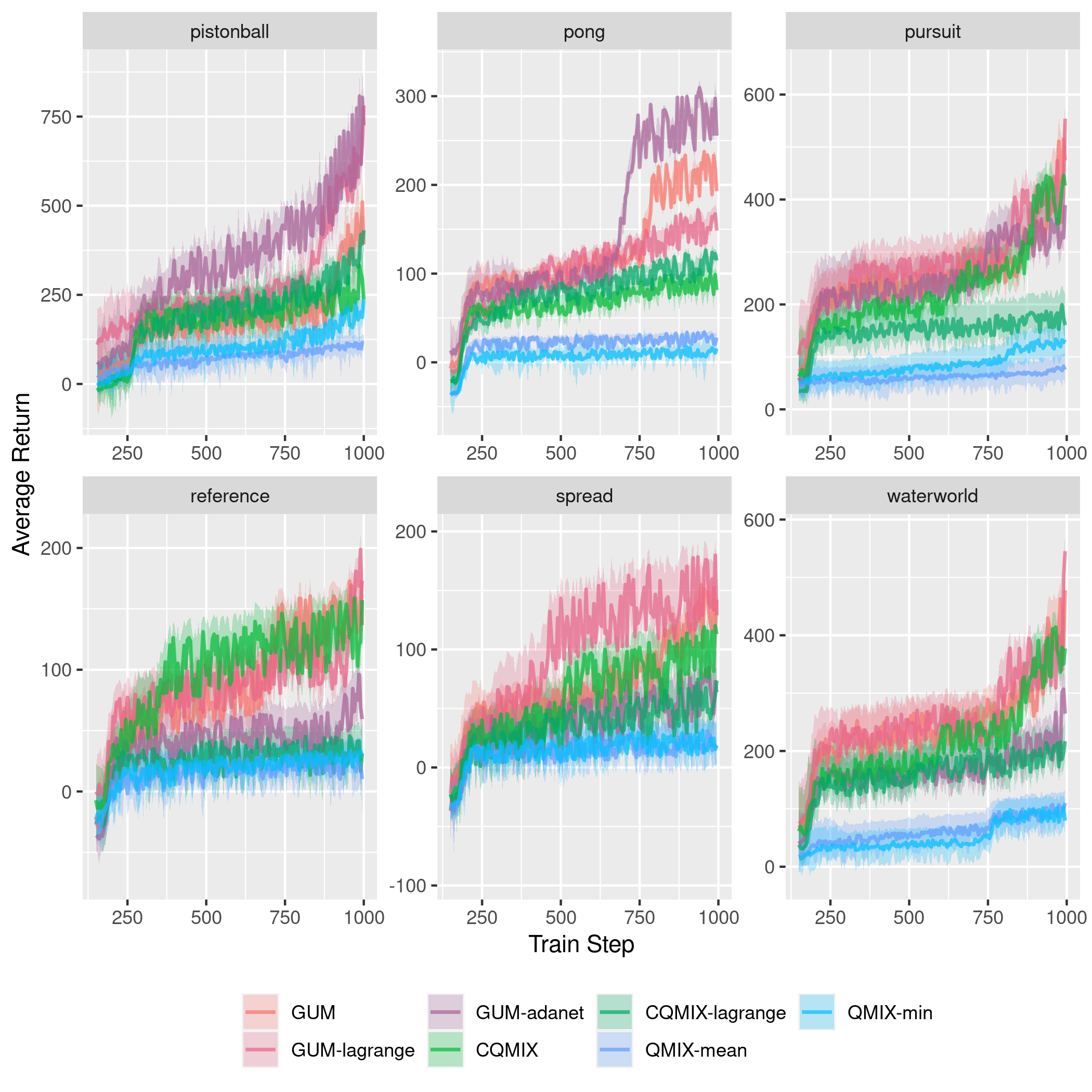} 
\includegraphics[width=0.45\textwidth]{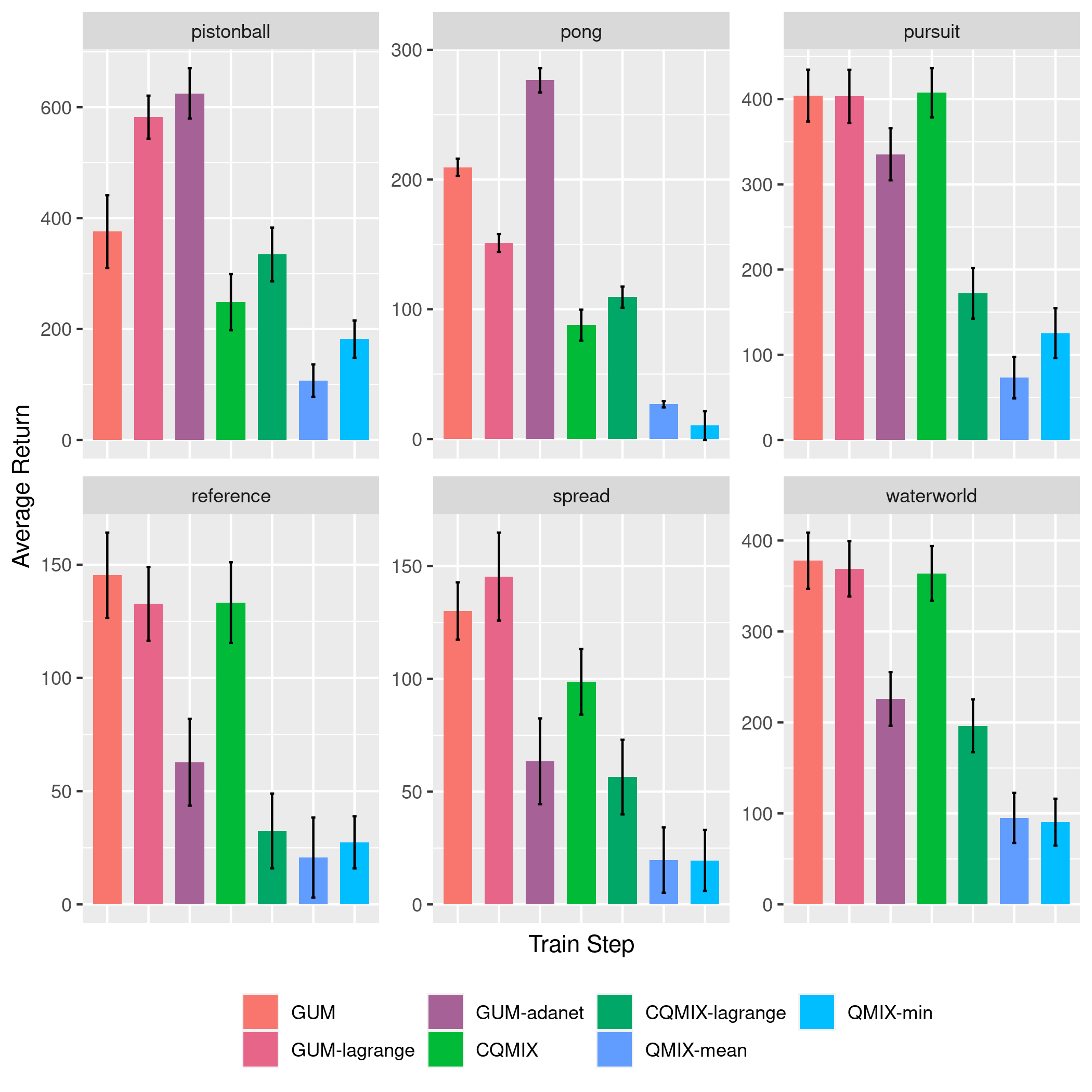}
\end{tabular}
  \end{center}
\caption{Performance over a variety of benchmark environments as part of our ablation studies for GUM algorithm. The bar graph reports the mean over the last 10 evaluation steps, where each evaluation point is average over 1000 episodes. Error bars for plots represent the standard deviation over these rollouts. This demonstrates the effectiveness of the GUM algorithm, as well as justifying the theoretical results, in particular the overestimation present in the explicit approach (``GUM-adanet'').}
\label{fig:ablations}
\end{figure}

\label{ablations}

In this section, we explore the effects of different variations of our algorithm on the performance of the agent in order to determine which algorithm generalizes best. We list the ``GUM'' algorithm as described in Section \ref{gum}, where the trade-off factor for the regularizer $\alpha$ is fixed, ``GUM-lagrange'' to be the GUM algorithm where the trade-off factor for the regularizer $\alpha$ is learned, ``GUM-adanet'' to use an explicit stacking with a fixed trade-off factor for the CQL regularizer. We contrast our approach CQMIX as described in Section \ref{cqmix}, which does not attempt to unmix the independent Q function and the mixing function, with a fixed trade-off factor for the regularizer and learned trade-off factor in algorithms labeled ``CQMIX'' and ``CQMIX-lagrange'' respectively. Finally to demonstrate the effectiveness of the algorithm is not related to the ensembling of agent Q policies with the mixing function, we use ensembling schemes involving the minimum or mean used in other RL approaches \cite{ddpg,bcq,bear} labeled ``QMIX-min'' and ``QMIX-mean'' respectively.

From the results, we observe that the GUM family of algorithms are generally the best performing algorithms. Surprisingly, explicitly stacking the networks together has strong performance in half of the environments (pistonball, pong, pursuit) but is inferior to a naive application of CQL to QMIX for the other environments (reference, spread, waterworld). For the other algorithms, it is clear that the naive application of CQL to QMIX and ensembling of QMIX algorithms is inferior to our GUM algorithm across a wide range of MARL tasks. This justifies our neural architecture choices and construction empirically. We hypothesize this is due to the dependent nature of the mixer hypernetwork and the independent Q-learning networks. In the same way target networks or Polyak averaging is used to minimise the update rate across the networks. Similar approaches can be used to learn the ensemble weighting, which appears to lead to degenerative cases. The inverse weighting scheme explicitly tempers the weights directly, based on the greedy weighting scheme. We leave exploring variations of avoiding this, as an exercise for future, perhaps through this mechanism, the networks may learn to unmix themselves without the need for inverse weighing scheme.



\subsection{Experimental Results}

\begin{figure}
  \begin{center}
\begin{tabular}{cc}
\includegraphics[width=0.45\textwidth]{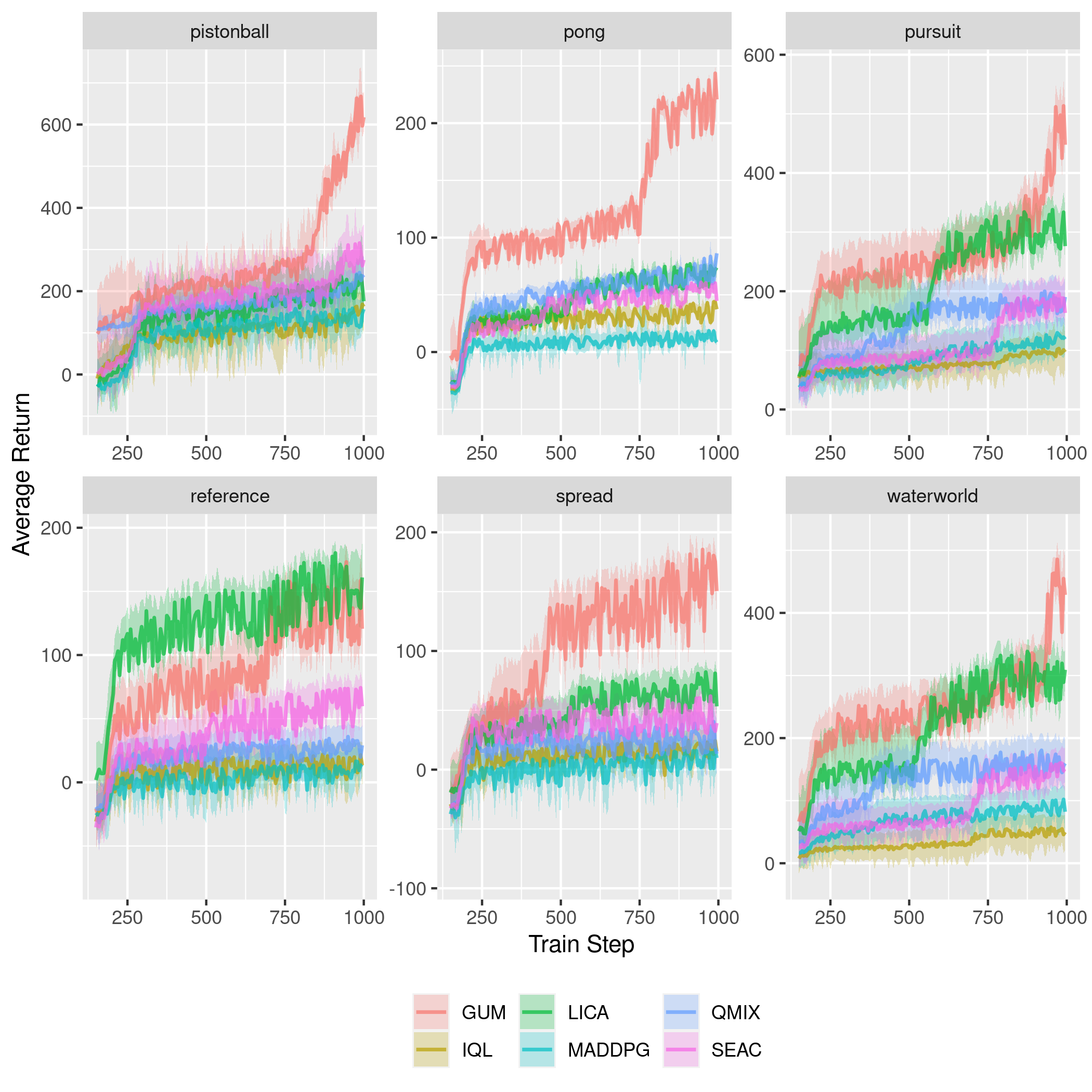}  \includegraphics[width=0.45\textwidth]{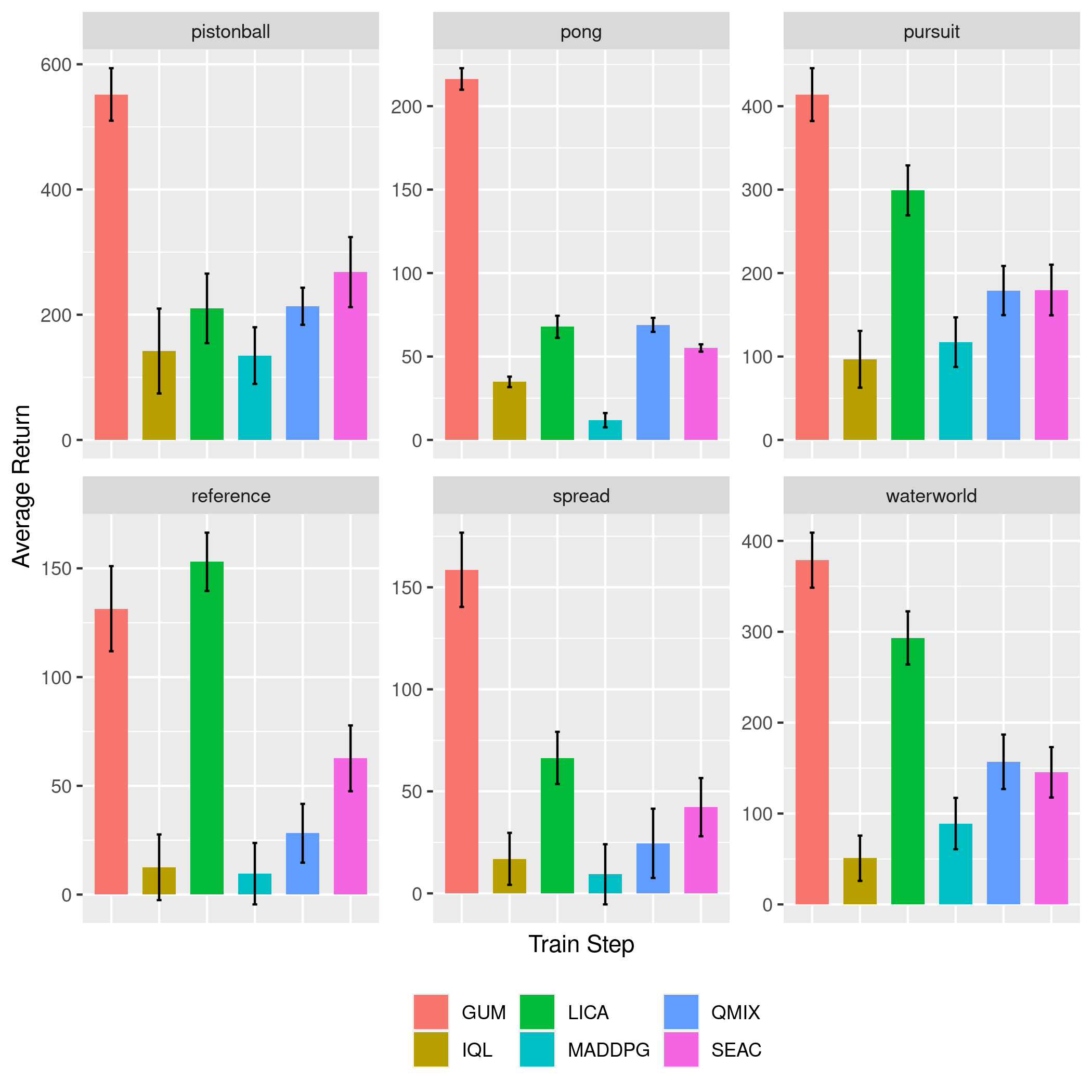}
\end{tabular}
  \end{center}
\caption{Performance over a variety of benchmark environments. The bar graph report the mean over the last 10 evaluation steps, where each evaluation point is average over 1000 episodes. Error bars for plots represent the standard deviation over these rollouts.}
\label{resall}
\end{figure}

We demonstrate that GUM is comparable to existing state of the art approaches, and performs well against models which use cooperation \cite{qmix} and entropy exploration \cite{lica,seac}. The GUM models leverage ``GUM-lagrange'' where the trade-off factors $\alpha$, $\alpha_{\text{mixer}}$ are learned implicitly. 

From the charts in Figure \ref{resall}, our approach is clearly superior or comparable with existing state of the art approaches in both Q-learning and actor-critic approaches for multi-agent reinforcement learning across a range of benchmark MARL tasks. The ability to train and effectively unmix the agent and the mixer network in a CTDE framework enables efficient training and representation in the GUM framework and suggests that GUM can effectively capture the nuances between independent agents and cooperative behavior. This is best observed when compared with LICA which captures implicit credit assignment in the actor-critic framework \cite{lica}. This suggests that our approach is competitive with actor-critic approaches in MARL whist maintaining the simplicity in neural network architecture and training as Q-learning approaches which may allow GUM to scale to large-scale problems effectively. 


\section{Conclusion and Future Work}

In this paper we explored Greedy UnMix (GUM), which is a dynamic ensemble technique for MARL in CTDE framework. We have explored and justified the construction of our network empirically through our experiments, and theoretically through linkages with frameworks for neural network architecture search. Being able to learn the effects of joint or individual policies is essential in learning underlying concepts effectively in cooperative multi-agent RL problems. GUM simplifies the joint state action space in the multi-agent setting through learning Q functions in a conservative fashion, and through weighing the trade-off between independent policies and mixer networks either implicitly or explicitly which leads to more purposeful coordination. Promising empirical results on a wide range of MARL benchmark environments and the simplicity of MARL Q-learning frameworks demonstrate the effectiveness of the algorithm in practice. For future work, we aim to extend this approach to the actor-critic class of algorithms and explore further theoretical properties of our architecture selection process in order to address effective ways to train MARL algorithms.


\bibliography{acml20}
\bibliographystyle{plain}

\end{document}